\documentclass{article}

\usepackage[preprint]{nips_2018}

\usepackage[utf8]{inputenc} 
\usepackage[T1]{fontenc}    
\usepackage{hyperref}       
\usepackage{url}            
\usepackage{booktabs}       
\usepackage{amsfonts}       
\usepackage{nicefrac}       
\usepackage{microtype}      
\usepackage{times} 
\usepackage{helvet} 
\usepackage{courier} 
\usepackage{graphicx} 
\usepackage{subfigure}
\usepackage{algorithm}  
\usepackage{algorithmicx}  
\usepackage{algpseudocode}

\title{Generative Model for Material Experiments \\ Based on Prior Knowledge and Attention Mechanism}

\author{
  Mincong Luo \\
  China Institute of Atomic Energy\\
  Peking,China \\
  \texttt{luomincentos@ciae.ac.cn} \\
  \And
  X. He \\
  China Institute of Atomic Energy\\
  Peking,China \\
  \texttt{xhe70@ciae.ac.cn}
  \And
  Li Liu \\
  China Institute of Atomic Energy\\
  Peking,China \\
  \texttt{liliu92@ciae.ac.cn} \\
}

\begin{document}

\maketitle

\begin{abstract}
Material irradiation experiment is dangerous and complex, thus it requires those with a vast advanced expertise to process the images and data manually. In this paper, we propose a generative adversarial model based on prior knowledge and attention mechanism to achieve the generation of irradiated material images (data-to-image model), and a prediction model for corresponding industrial performance (image-to-data model). With the proposed models, researchers can skip the dangerous and complex irradiation experiments and obtain the irradiation images and industrial performance parameters directly by inputing some experimental parameters only. We also introduce a new dataset ISMD which contains 22000 irradiated images with 22,143 sets of corresponding parameters. Our model achieved high quality results by compared with several baseline models. The evaluation and detailed analysis are also performed.

\end{abstract}

\section{introduction}
In recent years, significant progress has been made in the development of deep learning and generation models \cite{Kingma2013Auto} \cite{Goodfellow2014Generative} \cite{Mirza2014Conditional}. However, they are far less used in natural sciences than in services or the arts, experimental images from natural sciences have rich scientific connotations that can be analyzed using deep learning and generating models (e.g. medical images, fluid experiments, materials experiments etc. \cite{Kisilev2015From} \cite{Jing2017On} \cite{Alom2018Recurrent} \cite{Li2018Automated}).

Most of the researches on analyzing experimental images of natural sciences are based on feature extraction and end-to-end mapping to obtain valuable experimental target parameters \cite{Li2018Automated} \cite{Cire2013Mitosis} \cite{Jing2017On}, which can be called as \textbf{image-to-data} task. And most of these researches are based on convolutional neural networks(CNN) and semantic segmentation \cite{Ronneberger2015U} \cite{Menze2010A}. Although impressive results have been achieved, none of these researches have built a \textbf{data-to-image} model.


This research is based on the material irradiation experiment, which mainly focus on the material structural and performance changes after being irradiated. Researchers observe the structural changes by irradiated material images from electron microscope, and then confirm the performance changes by various mechanical experiments. 

We propose a generation model based on prior knowledge and attention mechanism to generate images of the experimental results. The overall architecture of the model is shown in Fig.3 .First, we propose an embedding model $fc(m)=Ph_m$ for the molecular composition of materials, so that we can get the feature vector $C^i_m$ for each material composition $i$, which is called \textbf{molecule2vec}.Then we can embed the feature vector $C^i_m$ of the material composition as a priori knowledge into the latent variable sampling of the generation model(see Fig.3 for details).For the irradiated material images, the swelling cavities distribution $H_v$ is the most important feature \cite{Ehrlich1981Irradiation}, so an attention mechanism is introduced to make the model focus on the section with swelling cavities in the material images(see Fig.1).The image generated by the feedforward propagation of the model $X'_{img} = G(D_d, D_c, z)$ can be encoded by the encoder network to be transformed into the corresponding swelling cavities distribution $H'_v$, we construct the loss term $\mathcal{L}_{ H_v} = || H'_v - H_v ||^2$ to measure the distance between the real distribution of the swelling cavities and the feature of the generated image , and then the attention mechanism is optimized by gradient descent.In the end, we can obtain high-quality images of irradiated materials. In addition, we pre-trained the image-to-data network $Pred(X_{img})=D_r$ to predict the performance parameters $D_r$ of the material through the corresponding experimental image $X_{img}$, which makes it possible to predict the material's performance parameters $D'_r$ by generating image $X'_{img}$ by $Pred(X'_{img})=D'_r$.


\textbf{our main contributions:}
\begin{itemize}
\item \textbf{data-to-image material images generation model:}generation model $P(X_{img}|D_d, D_c,z)$ is built based on prior knowledge and attention mechanism. 
\item \textbf{image-to-data material performance prediction model:} using the images $X_{img}$ and the context association method of CNN+BiLSTM, the network $P(D_r|X_{img}, C_m)$ is established to predict the performance parameters $D_r$ by the image $X_{img}$.
\item \textbf{The experimental dataset ISMD for irradiated materials:} the images with corresponding data from large number of irradiation experiments and manual annotations.
\end{itemize}

\begin{figure}   
\centering   
\subfigure[Overview]{     
\label{fig:subfig:a}
\includegraphics[scale=0.3]{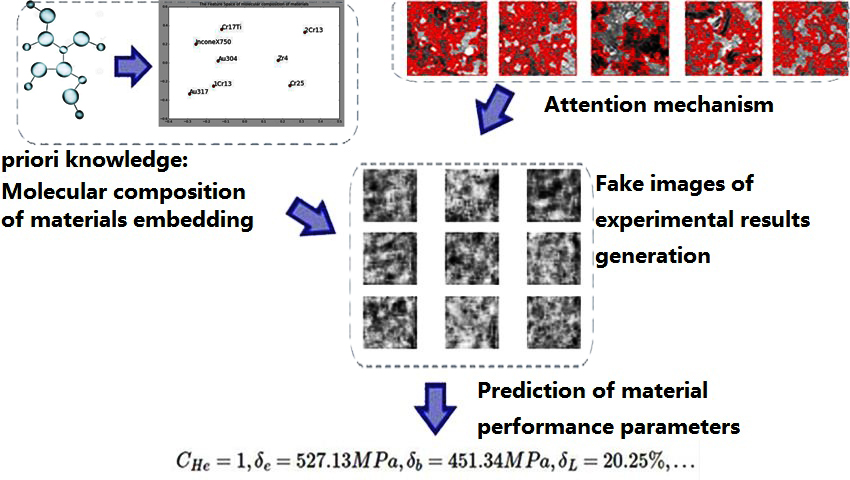}   }   
\subfigure[Representation Vector]{     
\label{fig:subfig:b}
\includegraphics[scale=0.22]{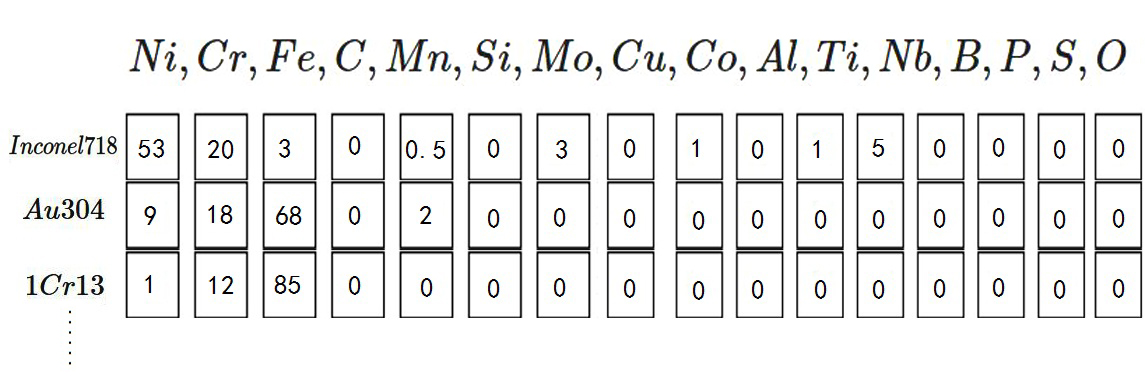}   }   
\caption{ \textbf{(a)} :\textbf{data-to-image model and image-to-data model}. In the data-to-image task, model generate images based on prior knowledge and attention mechanism. In the image-to-data task, images generated are used to predict the performance parameters. \textbf{(b)}: For different alloy materials, the proportion of the elements in their molecular compositions is calculated, which is used by element representation vector $m$. }   
\label{fig:subfig}
\end{figure}

\section{Related Work}
Our research is related to deep generation models for image generation, image information mining in the natural sciences, and irradiated material science.

\textbf{Deep generation model for image generation.} The caption-to-image generation model \cite{Gregor2015DRAW} \cite{Zhang2016StackGAN} \cite{Mansimov2015Generating} is the majority, most of these models will first introduce a semantic capture module to extract semantic features, then generate images through RNN \cite{Mansimov2015Generating}, stackGAN \cite{Zhang2016StackGAN}, etc. In addition, enhancing the connection between semantics and images by attention mechanism is also popular \cite{Zhang2017StackGAN}.

\textbf{Image information mining in the natural sciences.} Medicine, biology, materials and other fields have recently introduced deep learning models to mine the connotation information of experimental images \cite{Jung2017Iterative} \cite{Ronneberger2015U} \cite{Li2018Automated}. Some researches are inspired by natural language processing(NLP) using the context information association trick \cite{Ssm2017Auto}, which can be seen as an image-to-data process.

\textbf{Irradiated material science.} In recent years, deep learning methods have been introduced into material images processing \cite{Li2018Automated} \cite{Rovinelli2018Using}, which has replaced the traditional artificial technology analysis and achieved good results.

\section{Model}

Recall that our aim is to generate the corresponding experimental image $X'_{img}$ based on the mechanics and thermodynamic parameters $D_d$ of the material under the irradiation condition $D_c$ by the prior knowledge, the molecular composition feature vector $C_m$, and the attention mechanism. In addition, after generating experimental image $X'_{img}$, CNN+BiLSTM network $P(D'_r|X'_{img})$ is used to predict the performance parameters $D'_r$ for image $X'_{img}$.

\subsection{Embedding model for the molecular composition of materials}

Our dataset contains 14 kinds of alloy materials:

\[ C_m \in \{Inconel718, InconeX750,Zr1,Zr2,Zr4,Zr1Nb,Zr2.5Nb, \]
\[ 1Cr13,2Cr13,00Cr13Ni5Mo4,Au304,Au317,Cr17Ti ,Cr25\} \]

The molecular composition of these alloy materials is an important prior knowledge for images generation task, so embedding molecular composition features into the generation model is needed. Let $E=\{Al,Mg,Si,Cu,Fe,O...\}$ be the set of all the elements appeared. Let $m^i \in \mathbb{R}^{|E|}$ be the element percentage representation vector of the alloy material $i$. Inspired by the \textbf{word2vec} method \cite{Mikolov2013Efficient} for word embedding, a method for molecular composition of alloy materials feature extracting called \textbf{moleculars2vec} is proposed. $Ph_m = fc(m)$ is used to learn the thermodynamic properties $Ph_m$ from the alloy material by the element representation vector $m$ :

\[ Ph_m = fc(m) = Relu(W_m fc'(m) ) \]

where $fc(\cdot)$ represents a fully connected network, $fc'(\cdot)$ represents the network part of $fc(\cdot)$ without the last layer, $W_m \in \mathbb{R}^{ |E| \times d_{Ph } }$ is the weight matrix of the last layer, $d_{Ph}$ is the dimension of the thermodynamic property $Ph_m$. Since the weight matrix $W_m$ has the ability to represent both the molecular composition features and the properties of the alloy material, we take $C^i_m = W_m ^{i,:}$ as the feature vector of the material $i$.

\subsection{Generative model based on prior knowledge and attention mechanism}

\textbf{The prior distribution.} In the previous section we extracted the prior knowledge as vectors, that is, the features of the molecular composition of the alloy materials. To introduce it into the model, we define the distribution of the latent variable $P(z)$ as follow(prior distribution):

\[ z \sim \mathcal{N}(\mu(C_m) , \delta(C_m) ) \]
\[ \mu(C_m) = tanh(W_{\mu} C_m ) \]
\[ \delta(C_m) = exp(tanh(W_{\delta} C_m )) \]

where $W_{\mu} \in \mathbb{R}^{D \times dim(C_m) }$,$W_{\delta} \in \mathbb{R}^{D \times dim(C_m) }$ are the learnable parameters. Other similar methods introduce the distribution as latent variable dependencies are \cite{Mansimov2015Generating} \cite{Bachman2015Data}, and our model is optimized in performance after introducing a prior distribution. 

\textbf{The attention module.} The swelling cavities distribution $H_v$ of the irradiated material image is a statistics count vector for the distribution of different sizes, which is an important feature for material performance prediction \cite{Porollo2000Irradiation}. Therefore the model needs to allocate more attention to the section contains cavities, which requires attention mechanism to achieve. With this design goal in mind, we can set the attention module as an estimate of the corresponding degree of the cavities distribution $H_v$ and the image $X_{img}$, which is the soft attention mechanism $Attn(\cdot)$:

\[ \hat{X} = Attn(X_{conv},\hat{H}_v)= Relu(CNN(X_{img}) ,W_vH_v )\]

where $X_{conv} \in \mathbb{R}^{L \times L \times C}$ is a feature calculated by the forward propagation of convolutional layers, which with $C$ feature maps and with dimension of $L \times L$. $W_v \in \mathbb{R}^{|H_v| \times dim(X_{conv})}$ is a learnable parameter matrix that maps the swelling cavities distribution $H_v$ to the visual space. Finally, $\hat{X} = Attn( X_{conv}, \hat{H}_v)$ is inputed into the discriminator $D(\cdot)$ to complete feedforward propagation.

\subsection{Prediction Model For the performance of the materials}

After generating the image of the irradiated material, it is necessary to evaluate the various properties of the material under the condition $D_c$. A network is constructed to achieve the image-to-data task: $F_p : V \rightarrow \mathbb{R}^{dim(D_r)} $, where $V$ represents the visual space.

Before we design the structure of this performance evaluate network, let's revisit the two challenges in implementing this map:
(1) Irradiated material image $X_{img}$ and performance parameter $D_r$ are both highly abstract data forms. It is very difficult to establish mapping between two highly abstract data forms \cite{Jung2017Iterative}.
(2) The variables in the performance parameters $D_r$ are not independent to each other, which have mutual influence and physical connection.

To address the challenge 1, we introduce CNN to extract the visual features for the image $X_{img}$, and the molecular composition feature vector $C_m$ is introduced to change the model $P(D_r|X_{img})$ to $P(D_r|X_{img}, C_m)$; For the challenge 2, BiLSTM \cite{Schuster1996Bi} is used to learn the dependency of the variables in the performance parameters $D_r$. The prediction model is defined as:

\[ D^i_r = Relu(W_h[\overrightarrow{h_i},\overleftarrow{h_i}] + b_h)\]
\[ \overrightarrow{h_i} = \delta(W_xf \bar{X} + W_f h_{1:i-1} +b_f  ) \]
\[ \overleftarrow{h_i}  = \delta(W_xb \bar{X} + W_f h_{>i} +b_b  ) \]
\[ \bar{X} = Relu(W_x \hat{X}_{img} + W_m C_m + b ) \]

where $\hat{X}_{img} = CNN(X_{img})$ is the visual feature matrix extracted by convolutional neural networks(CNN), $\overrightarrow{h_i},\overleftarrow{h_i}$ are the forward and backward propagation hidden state vectors, which are concatenated together to predict the $i$-th variable in the performance parameters $D^i_r$.

\begin{figure}[H]
\centering
\includegraphics[scale=0.2]{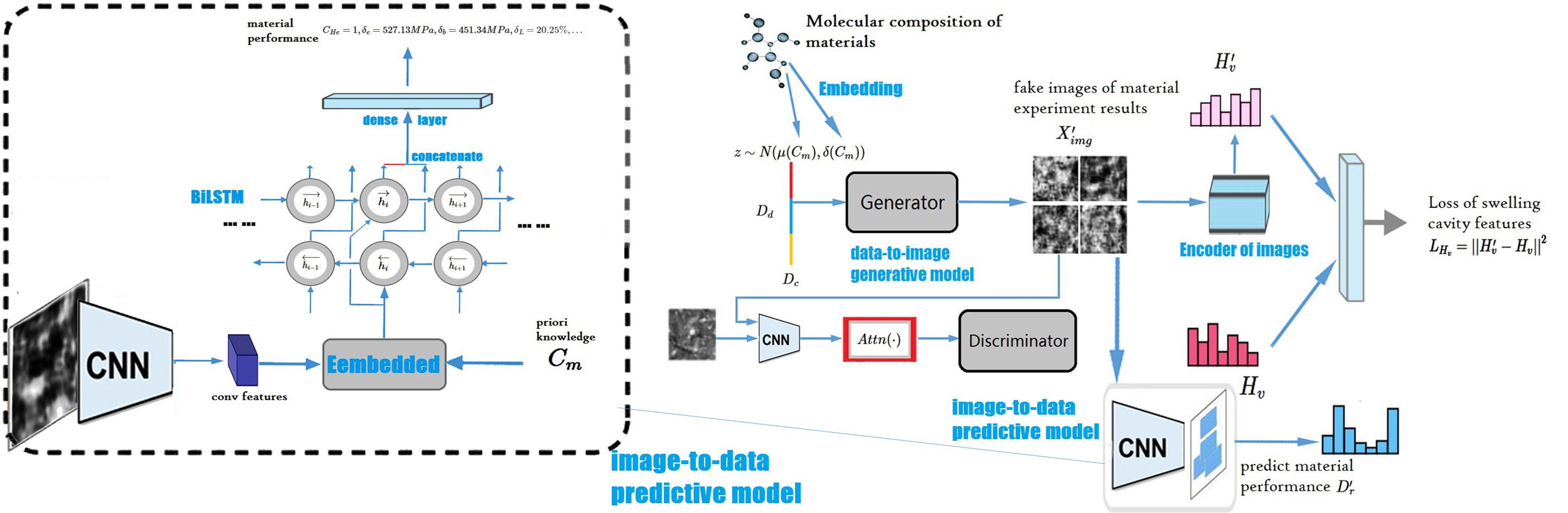}
\caption{The data-to-image model based on prior knowledge and attention mechanism achieves the generation of irradiated material images. Prior knowledge distribution is employed to embed the molecular composition of materials into the generative model. The attention mechanism is utilized to generate images with physical connotations(swelling cavities distribution). In the image-to-data task, CNN+BiLSTM is used to extract the features and learn the dependency of the performance variables.}
\end{figure}

\subsection{Learning}

In order to generate real images with physical connotations, the learning loss consists of two parts: GAN loss $\mathcal{L}_{GAN}$(prompting the model to generate real images) and swelling cavities feature loss $\mathcal{L}_{H_v}$(prompting the model to generate images with physical connotations, adjusting the parameters of attention mechanism by backpropagation). In order to calculate swelling cavities feature loss $\mathcal{L}_{H_v}$, We need to measure the corresponding degree of the image $X_{img}$ and the swelling cavities feature $H_v$, Which requires encoding the image.

\textbf{The image encoder:} image encoder maps the image $X_{img}$ to the space of the swelling cavities feature space by the convolutional neural network(CNN). The middle layer of the CNN can learn local features of the region in the image $X_{img}$. Finally, we map the CNN features of $X_{img}$ to the swelling cavities distribution features through the fully connected layers:$ \bar{H}_v =  En(X_{img}) = W fc(CNN(X_{img}) )$.

\textbf{The loss and learning steps:} the loss function consists of the loss of GAN $\mathcal{L}_{GAN}$ and the swelling cavities distribution features loss $\mathcal{L}_{H_v}$, and the final loss based on prior knowledge and attention mechanism is defined as $ \mathcal{L} = \mathcal{L}_{GAN} + \lambda \mathcal{L}_{H_v}$.

where $\lambda$ is a hyperparameter that balances two losses. The first loss represents the unconditional loss of GAN \cite{Xu2017AttnGAN}, and the second loss represents the conditional loss constrainted by the swelling cavities features. The loss $\mathcal{L}_{H_v}$ minimizes the difference between the cavities feature $H'_v$ of the generated image $X'_{img}$ and the true cavities feature $H_v$, defined as:$\mathcal{L}_{H_v} = || H'_v - H_v ||^2 = || En(X'_{img}) - H_v ||^2$.

The loss of GAN consists of the loss of the generator and the loss of the discriminator. We assign the swelling cavities feature loss $\mathcal{L}_{H_v}$ to the generator and the discriminator. The generator is trained to generate real images $X'_{img}$ with swelling cavities features(physical connotations), whose loss is defined as:

\[ \mathcal{L}_G = -\mathbb{E}_{X'_{img} \sim P_G }logD( X'_{img}, H_v)  + \frac{\lambda}{2} \mathcal{L}_{H_v} \]

The discriminator is trained to distinguish whether the input material image $X'_{img}$ is real or fake, whose loss is defined as:

\[ \mathcal{L}_D = -\mathbb{E}_{X_{img} \sim P_{data} }logD( X_{img}, H_v)-\mathbb{E}_{X'_{img} \sim P_G }log(1- D( X'_{img}, H_v) )  + \frac{\lambda}{2} \mathcal{L}_{H_v} \]

Finally, we can train the generator and discriminator alternately.

\begin{algorithm}[t]
\caption{The training algorithm using minibatch SGD with learning rate $\eta$}
\hspace*{0.02in} {\bf Input: minibatch images $X_{img}$; minibatch data $D_d,D_c,C_m,H_v$; minibatch size $S$;}  
\begin{algorithmic}[1]
\For{$n=1$ to $S$ do}
    \State $z \sim N(\mu(C_m) , \delta(C_m) )$;
    \State $X'_{img} \leftarrow G(z,D_d,D_c) $;
    \State $\mathcal{L}_r \leftarrow D(Attn(X_{img},H_v) ) $;
    \State $\mathcal{L}_f \leftarrow D(Attn(X'_{img},H_v) ) $;
    \State $\mathcal{L}_{H_v} \leftarrow ||En(X'_{img}) - H_v ||^2 $;
    \State $\mathcal{L}_D \leftarrow -log(\mathcal{L}_r) - log(1-\mathcal{L}_f) + \frac{\lambda}{2} \mathcal{L}_{H_v} $;
    \State $\theta_D \leftarrow \theta_D - \eta \bigtriangledown_{\theta_D} \mathcal{L}_D $;
    \State $\mathcal{L}_G \leftarrow -log(\mathcal{L}_f) + \frac{\lambda}{2} \mathcal{L}_{H_v} $;
    \State $\theta_G \leftarrow \theta_G - \eta \bigtriangledown_{\theta_G} \mathcal{L}_G  $;
\EndFor
\end{algorithmic}
\end{algorithm}

\subsection{Training details}
The training algorithm is implemented with the deep learning lib Keras. The networks are randomly intialized without any pre-training and is trained with decayed Adagrad and RMSprop. We train for a total of 2000 epochs and use a batch size of 10, a learning rate of $1 \times 10^{-5}$, and a weight decay rate of $1.6 \times 10^{-6}$ for the image generation task. The generated image size is $32 \times 32$. The molecule2vec embedding vectors are intialized as random vectors.

\section{Experiments}

Extensive experiments is carried out to evaluate the proposed model by comparing with multiple baseline models. First, we performed an experiment on image generation task and evaluate the ability of the model to generate experimental material images based on condition parameters(\textbf{data-to-image task}). Then the network $P(D'_r|X'_{img})$ predicting the performance parameters $D'_r$ by the image $X'_{img}$ is tested(\textbf{image-to-data task}), which is also compared with the theoretical models from computational materials science \cite{Ehrlich1981Irradiation} \cite{Boltax1978Design} \cite{Rest1999DART}.

\subsection{Dataset}
We introduce a dataset called \textbf{ISMD}(Irradiation Swelling Material Dataset) into this task. This dataset includes more than 22,000 images with size of $32 \times 32$ and with shooting scale of $100 nm$ from irradiation experiments, corresponding to the molecular compositions set $m$ of 14 alloys, material thermodynamics and mechanical properties set $D_d$, experimental condition parameters set $D_c$, swelling cavities distribution set $H_v$, and the performance parameters set $D_r$. The dataset is completed by data exported from the experimental devices, manually labeled, image preprocessed. More details about dataset ISMD can be found in Appendix.A;

\subsection{Baselines}

\textbf{Baseline models for image generation.}
\begin{itemize}
\item \textbf{Pure GAN:} the prior distribution and the attention mechanism removed from the proposed model.
\item \textbf{The prior distribution dropped:} only the prior distribution of the proposed model removed and is used to evaluate the value of prior distribution.
\item \textbf{The attention mechanism dropped:} only the attention mechanism of the proposed model removed and is used to evaluate the value of the attention mechanism.
\item \textbf{Variational Auto-Encoder:} To verify whether GAN is a better generation model for this task.
\end{itemize}

\textbf{Baseline model for materials performance prediction:} The theoretical model from computational material science is used to evaluate the prediction model in image-to-data task. 


\begin{figure}   
\centering   
\subfigure[data-to-image]{     
\label{fig:subfig:a}
\includegraphics[scale=0.23]{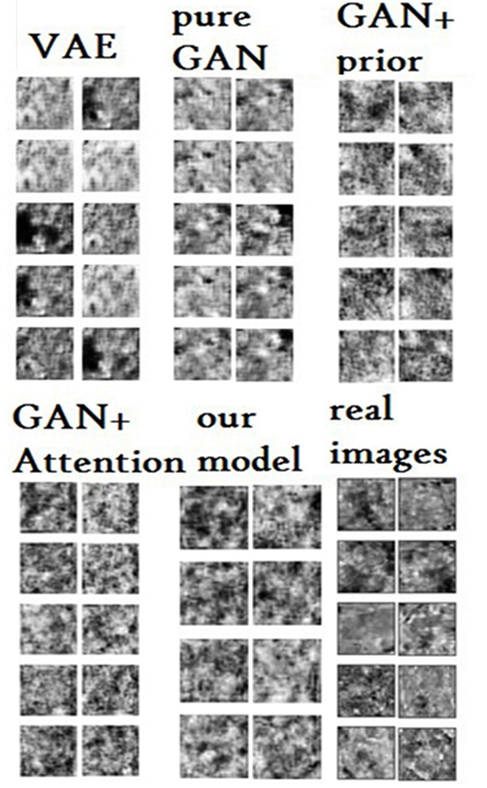}   }   
\subfigure[image-to-data]{     
\label{fig:subfig:b}
\includegraphics[scale=0.43]{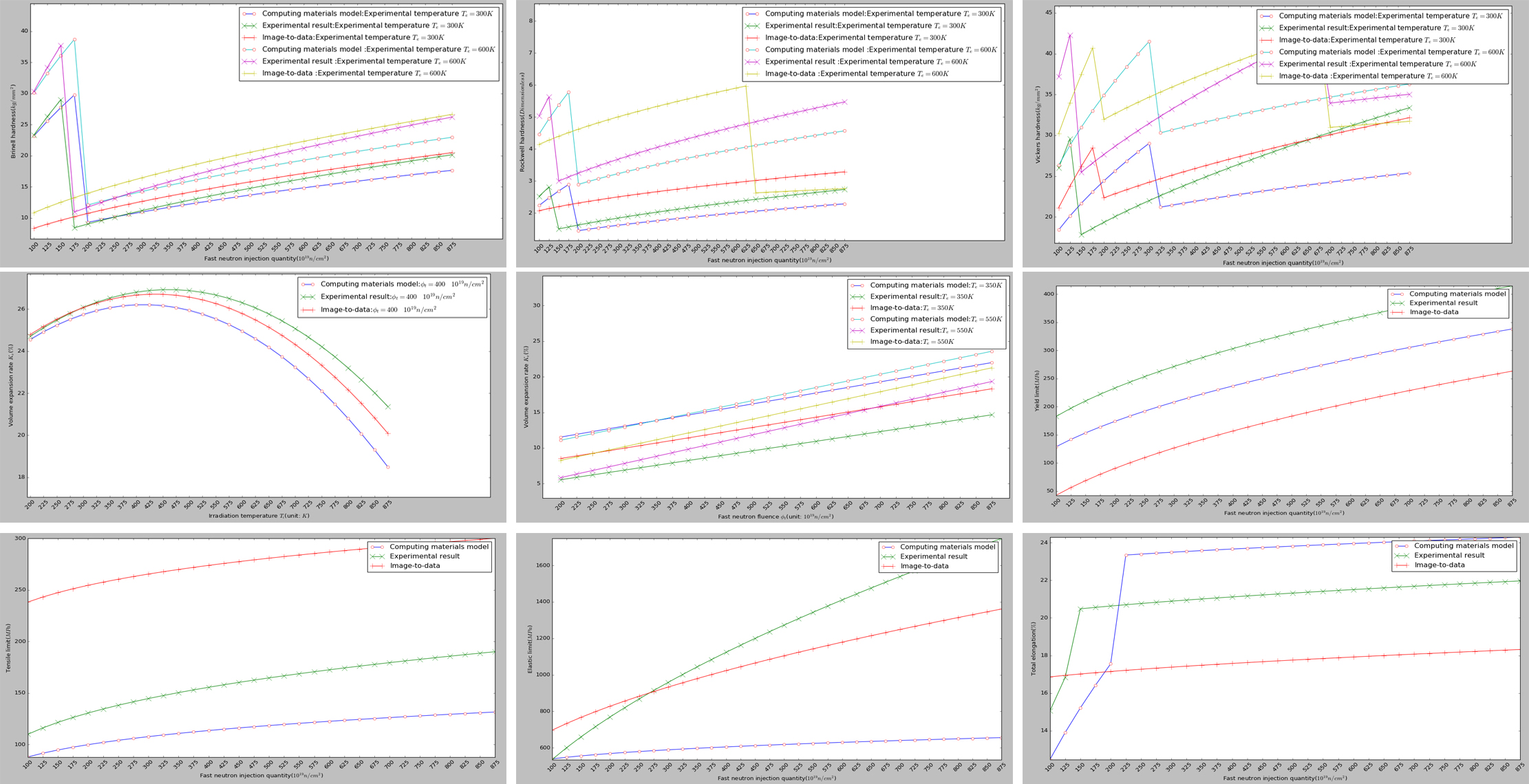}   }   
\caption{ \textbf{(a)}:The images generated by the proposed model and the baseline models. \textbf{(b)}:The materials performance prediction model compared to theoretical models and the experimental results.}   
\label{fig:subfig}
\end{figure}

\subsection{Evaluation Metrics}
\textbf{Evaluation metrics for image generation:} Inception score \cite{Salimans2016Improved}  \cite{Springenberg2015Unsupervised}is the evaluation method for generating images. The core idea of the method is: images contain meaningful objects should have a conditional label distribution $p(y|x)$ with low entropy, moreover, the model is expected to generate varied images, so the marginal $\int p(y|x' = G(z))dz$ should have high entropy.Combining these two requirements, the evaluation is defined as $exp(\mathbb{E}_{m} KL(p(X'_{img}|m) || p(X_{img})))$, where $m$ is the material representation vector.

\textbf{Evaluation metrics for materials performance prediction:} the material performance parameters from the performance evaluation experiments is $D^*_r$, the parameters calculated by the theoretical models \cite{Ehrlich1981Irradiation} \cite{Boltax1978Design} \cite{Rest1999DART} is $\hat{D}_r$, and which predicted by the image $X_{img}$ in the proposed model $P(D_r|X_{img}, C_m)$ is $D'_r$. Finally, the accuracy evaluation is defined as the RMSE score $S_{RMSE} =  \sqrt{ \frac{1}{m} \sum_i(D'^{(i)}_r - \hat{D^{(i)}_r})^2 } $.


\begin{minipage}{\textwidth}
 \begin{minipage}[t]{0.45\textwidth}
  \centering
     \makeatletter\def\@captype{table}\makeatother\caption{The data-to-image task.}
       \begin{tabular}{lclll}
            \toprule
            models &Inception score\\
            \midrule
            VAE       & 1.89 $\pm$ 0.05\\
            GAN       & 2.30 $\pm$ 0.07\\
            PR+GAN    & 2.40 $\pm$ 0.02\\
            Att+GAN   & 2.84 $\pm$ 0.07\\
            Our Model & \textbf{3.87 $\pm$ 0.08}\\
            Real Data & 7.11 $\pm$ 0.14\\
          \bottomrule
        \end{tabular}
  \end{minipage}
  \begin{minipage}[t]{0.45\textwidth}
   \centering
        \makeatletter\def\@captype{table}\makeatother\caption{The \textbf{accuracy evaluation} for materials performance prediction.}
        \begin{tabular}{lclll}
            \toprule
            prediction tasks & img-to-data & theoretical models\\
            \midrule
            $\delta_s$ & 1.96 & \textbf{1.80} \\
            $\delta_b$ & 1.92 & \textbf{1.63} \\
            $\delta_e$ & \textbf{1.89} & 1.91 \\
            $\delta_L$ & 1.98 & \textbf{1.55} \\
            $H_B$     & 2.19 & \textbf{1.75} \\
            $H_{RC}$  & 2.23 & \textbf{1.57} \\
            $H_V$     & \textbf{2.21} & 2.24 \\
            $K$       & \textbf{1.96} & 1.98 \\
          \bottomrule
        \end{tabular}
   \end{minipage}
\end{minipage}

\subsection{Results}

For the \textbf{data-to-image} task, the proposed model performs better on the ISMD dataset than other baseline models, achieving higher scores.For the \textbf{image-to-data} task, our model performs better on the prediction of some material properties compared to the theoretical model. In summary, the combination of the image-to-data model and computational material science models should be better.

\begin{figure}[H]
\centering
\includegraphics[scale=0.98]{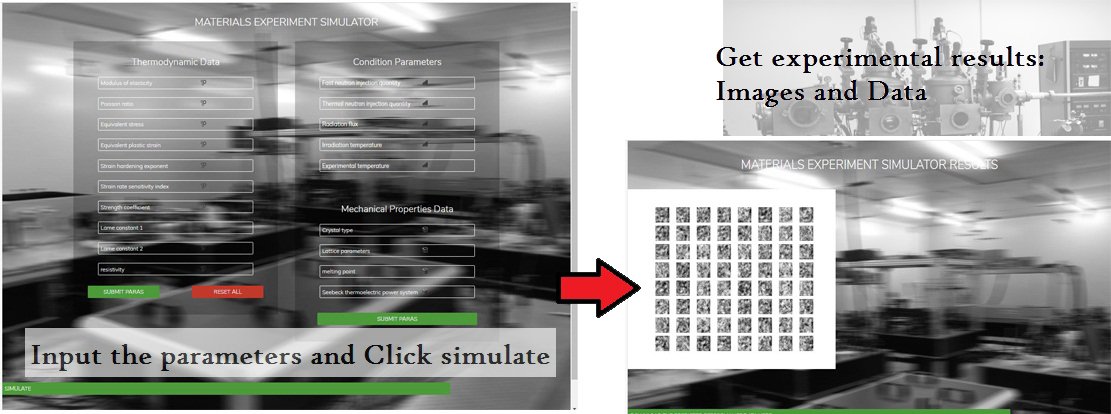}
\caption{We put the trained models on the server and developed a \textbf{Web App}.}
\end{figure}

\section{Conclusion}
In this work, we propose the data-to-image model based on prior knowledge and attention mechanism to achieve the generation of irradiated material images. In the image-to-data task, CNN and BiLSTM are used to extract the features and learn the dependency of the variables. In the future work, we plan to generate more controllable and diverse material images with physical connotations and predict the performance data to replace the experiments.

\appendix       
\section*{Appendix}

\begin{figure}[H]
\centering
\includegraphics[scale=0.4]{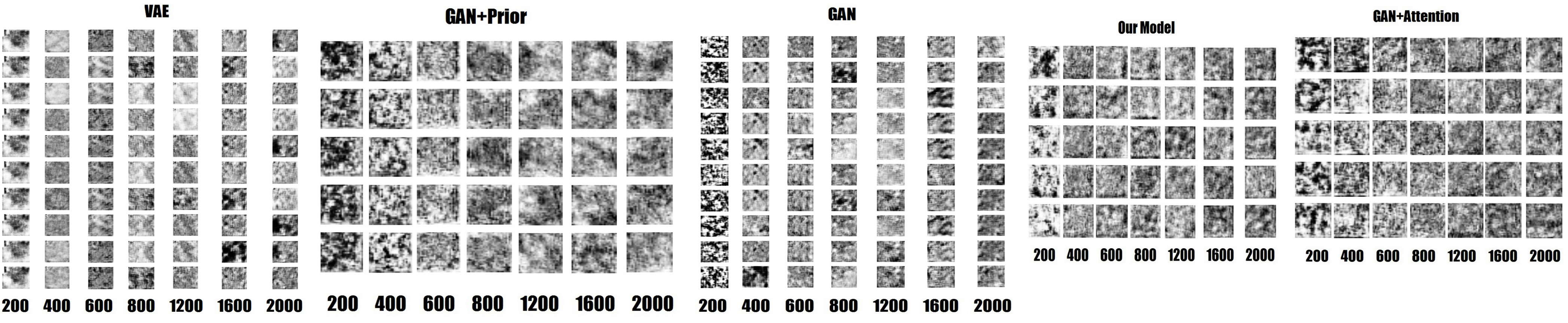}
\caption{ Images generation of the baseline models and proposed model from different epochs.}
\end{figure}

\subsection{material thermodynamics and mechanical properties set $D_d$}

\textbf{mechanical properties:} \\

$E$: modulus of elasticity (unit: $MPa$);  $\nu$: Poisson's ratio;  \\
$\bar{\delta}$: Equivalent stress (unit: $MPa$);  $\bar{\varepsilon}$: Equivalent plastic strain (unit: $MPa$);  \\
$n$: strain hardening index;  $m$: strain rate sensitivity index;  \\
$K$: intensity factor;  Lame constant $\lambda$;  \\
Lame constant $G$; \\

\textbf{thermodynamics:} \\

$C_t$: crystal type;    $C_d$: lattice parameter (unit: $nm$); \\
$T_m$: melting point (unit: $K$);    $\rho$: theoretical density (unit: $g/cm^3$); \\
$\hat{V}$: Thermal expansion coefficient (unit: $10^{-6} \cdot K^{-1}$);    $C_h$: Thermal conductivity (unit: $W \cdot m^{-1} \cdot C^{-1}$); \\
$H_c$: heat capacity (unit: $KJ/mol \cdot K$);    $E_h$: hot (unit: $KJ/mol$); \\
$C_e$: Seebeck temperature difference electromotive force factor (unit: $uV/K$);    $C_R$: resistivity (unit: $10^3 \cdot \Omega \cdot cm$); \\

\subsection{experimental condition parameters set $D_d$}

Fast neutron injection volume $\phi_f$(unit: $10^{19}n/cm^2$);   Thermal neutron injection amount $\phi_t$(unit: $10^{19}n/cm^2$); \\
Irradiation flux $\phi_i$(unit: $10^{19}n/cm^2$);   Irradiation temperature $T_i$(unit: $K$); \\
Experimental temperature $T_e$(unit: $K$); \\


\begin{figure}
\begin{minipage}[H]{0.5\linewidth}
\centering
\includegraphics[scale=0.23]{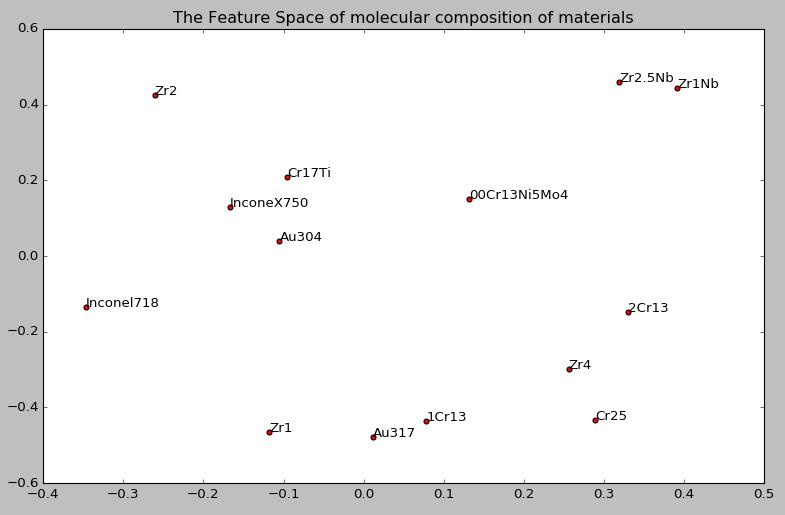}
\label{fig:side:a}
\end{minipage}%
\begin{minipage}[H]{0.5\linewidth}
\centering
\includegraphics[scale=0.6]{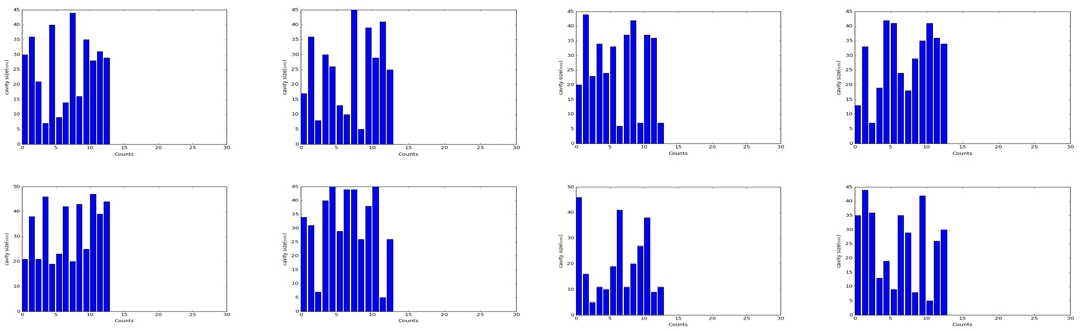}
\caption{\textbf{Left}: We visualize the distribution of the molecular composition feature vector $C_m$ in two-dimensional space.\textbf{Right}: The swelling cavities distribution $H_v$.}
\label{fig:side:b}
\end{minipage}
\end{figure}

\subsection{performance parameters set $D_r$}

Yield limit $\delta_s$(unit: $MPa$);    Stretch limit $\delta_b$(unit: $MPa$); \\
Elastic limit $\delta_e$(unit: $MPa$);    The total extension rate is $\delta_L$(unit: $\%$); \\
Brinell hardness $H_B$(unit: $kg/mm^2$);    Rockwell hardness $H_{RC}$(unit: $mm$) (dimensionless: HR*T); \\
Vickers hardness $H_{V}$(unit: $kg/mm^2$);    Volume expansion rate $K_v$(unit: $\%$); \\
Irradiation growth rate $K_L$(unit: $\%$);    Fracture toughness $K_{Ic}$(unit: $MPa/m^{1/2}$); \\
Creep performance $\delta_t$(unit: $MPa$);    Crisp feature $C_{He} \in \{0,1\}$; \\





\bibliography{ref}

\begin{thebibliography}{27}
\providecommand{\natexlab}[1]{#1}
\providecommand{\url}[1]{\texttt{#1}}
\expandafter\ifx\csname urlstyle\endcsname\relax
  \providecommand{\doi}[1]{doi: #1}\else
  \providecommand{\doi}{doi: \begingroup \urlstyle{rm}\Url}\fi

\bibitem[Alom et~al.(2018)Alom, Hasan, Yakopcic, Taha, and
  Asari]{Alom2018Recurrent}
Md~Zahangir Alom, Mahmudul Hasan, Chris Yakopcic, Tarek~M. Taha, and Vijayan~K.
  Asari.
\newblock Recurrent residual convolutional neural network based on u-net
  (r2u-net) for medical image segmentation.
\newblock \emph{IEEE Conf}, 2018.

\bibitem[Bachman \& Precup(2015)Bachman and Precup]{Bachman2015Data}
Philip Bachman and Doina Precup.
\newblock Data generation as sequential decision making.
\newblock \emph{Computer Science}, pp.\  3249--3257, 2015.

\bibitem[Boltax et~al.(1978)Boltax, Foster, Kalinowski, and
  Swenson]{Boltax1978Design}
A.~Boltax, J.~P. Foster, J.~E. Kalinowski, and D.~C. Swenson.
\newblock Design applications of irradiation creep and swelling data.
\newblock \emph{Trans. Am. Nucl. Soc.; (United States)}, 28, 1978.

\bibitem[Cireşan et~al.(2013)Cireşan, Giusti, Gambardella, and
  Schmidhuber]{Cire2013Mitosis}
D.~C. Cireşan, A~Giusti, L.~M. Gambardella, and J~Schmidhuber.
\newblock Mitosis detection in breast cancer histology images with deep neural
  networks.
\newblock In \emph{International Conference on Medical Image Computing and
  Computer-Assisted Intervention}, pp.\  411--8, 2013.

\bibitem[Ehrlich(1981)]{Ehrlich1981Irradiation}
Karl Ehrlich.
\newblock Irradiation creep and interrelation with swelling in austenitic
  stainless steels.
\newblock \emph{Journal of Nuclear Materials}, 100\penalty0 (1):\penalty0
  149--166, 1981.

\bibitem[Goodfellow et~al.(2014)Goodfellow, Pouget-Abadie, Mirza, Xu,
  Warde-Farley, Ozair, Courville, and Bengio]{Goodfellow2014Generative}
Ian~J. Goodfellow, Jean Pouget-Abadie, Mehdi Mirza, Bing Xu, David
  Warde-Farley, Sherjil Ozair, Aaron Courville, and Yoshua Bengio.
\newblock Generative adversarial nets.
\newblock In \emph{International Conference on Neural Information Processing
  Systems}, pp.\  2672--2680, 2014.

\bibitem[Gregor et~al.(2015)Gregor, Danihelka, Graves, Rezende, and
  Wierstra]{Gregor2015DRAW}
Karol Gregor, Ivo Danihelka, Alex Graves, Danilo~Jimenez Rezende, and Daan
  Wierstra.
\newblock Draw: a recurrent neural network for image generation.
\newblock \emph{Computer Science}, pp.\  1462--1471, 2015.

\bibitem[Jing et~al.(2017)Jing, Xie, and Xing]{Jing2017On}
Baoyu Jing, Pengtao Xie, and Eric Xing.
\newblock On the automatic generation of medical imaging reports.
\newblock \emph{ICML}, 2017.

\bibitem[Jung et~al.(2017)Jung, Hak, and Yong]{Jung2017Iterative}
Kim~Uk Jung, Kim~Gu Hak, and Ro~Man Yong.
\newblock Iterative deep convolutional encoder-decoder network for medical
  image segmentation.
\newblock \emph{IEEE Conf}, 2017.

\bibitem[Kingma \& Welling(2013)Kingma and Welling]{Kingma2013Auto}
Diederik~P Kingma and Max Welling.
\newblock Auto-encoding variational bayes.
\newblock \emph{Arxiv}, 2013.

\bibitem[Kisilev et~al.(2015)Kisilev, Walach, Barkan, Ophir, Alpert, and
  Hashoul]{Kisilev2015From}
P.~Kisilev, E.~Walach, E.~Barkan, B.~Ophir, S.~Alpert, and S.~Y. Hashoul.
\newblock From medical image to automatic medical report generation.
\newblock \emph{Ibm Journal of Research and Development}, 59\penalty0
  (2/3):\penalty0 2:1--2:7, 2015.

\bibitem[Li et~al.(2018)Li, Field, and Morgan]{Li2018Automated}
Wei Li, Kevin~G. Field, and Dane Morgan.
\newblock Automated defect analysis in electron microscopic images.
\newblock \emph{Arxiv}, 2018.

\bibitem[Mansimov et~al.(2015)Mansimov, Parisotto, Ba, and
  Salakhutdinov]{Mansimov2015Generating}
Elman Mansimov, Emilio Parisotto, Jimmy~Lei Ba, and Ruslan Salakhutdinov.
\newblock Generating images from captions with attention.
\newblock \emph{Computer Science}, 2015.

\bibitem[Menze et~al.(2010)Menze, Van, Lashkari, Weber, Ayache, and
  Golland]{Menze2010A}
B.~H. Menze, Leemput~K Van, D~Lashkari, M.~A. Weber, N~Ayache, and P~Golland.
\newblock A generative model for brain tumor segmentation in multi-modal
  images.
\newblock \emph{Ibm Journal of Research and Development}, 13\penalty0
  (2):\penalty0 151--9, 2010.

\bibitem[Mikolov et~al.(2013)Mikolov, Chen, Corrado, and
  Dean]{Mikolov2013Efficient}
Tomas Mikolov, Kai Chen, Greg Corrado, and Jeffrey Dean.
\newblock Efficient estimation of word representations in vector space.
\newblock \emph{Computer Science}, 2013.

\bibitem[Mirza \& Osindero(2014)Mirza and Osindero]{Mirza2014Conditional}
Mehdi Mirza and Simon Osindero.
\newblock Conditional generative adversarial nets.
\newblock \emph{Computer Science}, pp.\  2672--2680, 2014.

\bibitem[Porollo et~al.(2000)Porollo, Vorobjev, Konobeev, Budylkin, Mironova,
  Garner, Porollo, Vorobjev, Konobeev, and Budylkin]{Porollo2000Irradiation}
S.~I. Porollo, A.~N. Vorobjev, Yu.~V. Konobeev, N.~I. Budylkin, E.~G. Mironova,
  F.~A. Garner, S.~I. Porollo, A.~N. Vorobjev, Yu.~V. Konobeev, and N.~I.
  Budylkin.
\newblock Irradiation creep and stress-affected swelling in austenitic
  stainless steel 16cr-15ni-3mo-nb-b irradiated in the bn-350 reactor.
\newblock \emph{Aaps Pharmscitech}, 13\penalty0 (4):\penalty0 1386--1395, 2000.

\bibitem[Rest \& Hofman(1999)Rest and Hofman]{Rest1999DART}
J.~Rest and G.~L. Hofman.
\newblock Dart model for irradiation-induced swelling of uranium silicide
  dispersion fuel elements.
\newblock \emph{Nuclear Technology}, 126\penalty0 (1):\penalty0 88--101, 1999.

\bibitem[Ronneberger et~al.(2015)Ronneberger, Fischer, and
  Brox]{Ronneberger2015U}
Olaf Ronneberger, Philipp Fischer, and Thomas Brox.
\newblock \emph{U-Net: Convolutional Networks for Biomedical Image
  Segmentation}.
\newblock Springer International Publishing, 2015.

\bibitem[Rovinelli(2018)]{Rovinelli2018Using}
A~Rovinelli.
\newblock Using machine learning and a data-driven approach to identify the
  small fatigue crack driving force in polycrystalline materials.
\newblock \emph{Arxiv}, 2018.

\bibitem[Salimans et~al.(2016)Salimans, Goodfellow, Zaremba, Cheung, Radford,
  and Chen]{Salimans2016Improved}
Tim Salimans, Ian Goodfellow, Wojciech Zaremba, Vicki Cheung, Alec Radford, and
  Xi~Chen.
\newblock Improved techniques for training gans.
\newblock \emph{Computer Science}, 2016.

\bibitem[Schuster(1996)]{Schuster1996Bi}
Mike Schuster.
\newblock Bi-directional recurrent neural networks for speech recognition.
\newblock \emph{Ieice Technical Report Speech}, 96:\penalty0 7--12, 1996.

\bibitem[Springenberg(2015)]{Springenberg2015Unsupervised}
Jost~Tobias Springenberg.
\newblock Unsupervised and semi-supervised learning with categorical generative
  adversarial networks.
\newblock \emph{Computer Science}, 2015.

\bibitem[Ssm et~al.(2017)Ssm, Erdogmus, and Gholipour]{Ssm2017Auto}
Salehi Ssm, D~Erdogmus, and A~Gholipour.
\newblock Auto-context convolutional neural network (auto-net) for brain
  extraction in magnetic resonance imaging.
\newblock \emph{IEEE Transactions on Medical Imaging}, PP\penalty0
  (99):\penalty0 1--1, 2017.

\bibitem[Xu et~al.(2017)Xu, Zhang, Huang, Zhang, Gan, Huang, and
  He]{Xu2017AttnGAN}
Tao Xu, Pengchuan Zhang, Qiuyuan Huang, Han Zhang, Zhe Gan, Xiaolei Huang, and
  Xiaodong He.
\newblock Attngan: Fine-grained text to image generation with attentional
  generative adversarial networks.
\newblock \emph{Arxiv}, 2017.

\bibitem[Zhang et~al.(2016)Zhang, Xu, and Li]{Zhang2016StackGAN}
Han Zhang, Tao Xu, and Hongsheng Li.
\newblock Stackgan: Text to photo-realistic image synthesis with stacked
  generative adversarial networks.
\newblock \emph{Computer Science}, pp.\  5908--5916, 2016.

\bibitem[Zhang et~al.(2017)Zhang, Xu, Li, Zhang, Wang, Huang, and
  Metaxas]{Zhang2017StackGAN}
Han Zhang, Tao Xu, Hongsheng Li, Shaoting Zhang, Xiaogang Wang, Xiaolei Huang,
  and Dimitris~N. Metaxas.
\newblock Stackgan++: Realistic image synthesis with stacked generative
  adversarial networks.
\newblock \emph{IEEE Transactions on Pattern Analysis and Machine
  Intelligence}, PP\penalty0 (99), 2017.

\end{thebibliography}
\bibliographystyle{iclr2019_conference}

\end{document}